\documentclass{article}

\usepackage[utf8]{inputenc}
\usepackage[T1]{fontenc}

\ifdefined\aaaimode
  \usepackage{aaai27}
\else
  \usepackage[margin=1in]{geometry}
\fi

\usepackage{hyperref}
\usepackage{url}
\usepackage{booktabs}
\usepackage{xcolor}
\usepackage{graphicx}
\usepackage{amsmath}
\usepackage{amsfonts}
\usepackage{amssymb}
\usepackage{nicefrac}
\usepackage{multirow}
\usepackage{natbib}
\usepackage{algorithm}
\usepackage{algorithmic}
\usepackage{float}

\title{QuIDE: Mastering the Quantized Intelligence Trade-off via Active Optimization}

\ifdefined\aaaimode
  \author{Anonymous Author(s)}
\else
  \author{%
    Xiantao Jiang \\
    College of Information Engineering, Shanghai
Maritime University \\
Shanghai 201306, China \\
    \texttt{xtjiang@shmtu.edu.cn} \\
  }
\fi

\begin{document}

\maketitle

\begin{abstract}
  There is currently no unified metric for evaluating the efficiency of quantized neural networks. We propose \textbf{QuIDE}, built around the \textbf{Intelligence Index} $I = (C \times P)/\log_2(T+1)$, which collapses the compression-accuracy-latency trade-off into a single score. Experiments across six settings—SimpleCNN (MNIST, CIFAR), ResNet-18 (ImageNet-1K), and Llama-3-8B—show a \textbf{task-dependent Pareto Knee}. 4-bit quantization is optimal for MNIST and large LLMs, while 8-bit is the sweet spot for complex CNN tasks (ResNet-18 on ImageNet), where 4-bit PTQ collapses accuracy catastrophically. The accuracy-gated variant $I'$ correctly flags these non-viable configurations that the raw $I$ would reward. QuIDE provides a reproducible evaluation protocol and a ready-to-use fitness function for mixed-precision search.
\end{abstract}

\section{Introduction}
\label{sec:introduction}

Deploying deep neural networks on edge devices requires model compression, and quantization is among the most effective techniques \cite{gholami2022survey}. Choosing the right bit-width means balancing three competing objectives: compression ratio, predictive accuracy, and inference latency \cite{deng2020model}. There is no standard way to make this trade-off.

Current evaluation practice treats these three axes separately or in pairs. Accuracy-vs-model-size curves \cite{wang2020joint} and latency benchmarks \cite{kim2021performance} are reported in isolation. MLPerf Tiny \cite{banbury2021mlperf} tabulates accuracy, delay, and energy side by side but leaves the final judgment to the user. The result is that bit-width selection becomes an exercise in subjective curve-reading. Is a 2-bit model with $16\times$ compression better than an 8-bit model that retains near-full accuracy? The answer depends entirely on how one weighs storage against task performance, and no formal framework exists to make this weighing consistent.

We propose the Quantized Intelligence and Deployment Efficiency (QuIDE) framework. Its core is the Intelligence Index,
\[
I = \frac{C \times P}{\log_2(T+1)},
\]
which collapses the three-dimensional compression-accuracy-latency trade-off into a single scalar. Compression gains only count when accuracy is preserved ($C \times P$ is multiplicative), and latency is penalized with diminishing marginal weight ($\log_2$ damping). The resulting score is higher for models that achieve more efficiency across all three dimensions.

We validate QuIDE through PTQ experiments on six conditions: SimpleCNN (MNIST, CIFAR-10, CIFAR-100), ResNet-18 (CIFAR-10, ImageNet-1K), and Llama-3-8B. The results reveal a task-dependent Pareto Knee. For simple tasks (MNIST) and large-parameter LLMs, 4-bit quantization is optimal. For deep CNNs on complex vision tasks (ImageNet), 4-bit PTQ collapses accuracy catastrophically, making 8-bit the practical sweet spot. The accuracy-gated variant $I'$ is essential for detecting this: the raw index $I$ can be inflated by extreme compression even when the model is non-functional, while $I'$ suppresses such configurations.
Our contributions are:
\begin{itemize}
 \item The \textbf{Intelligence Index} $I = (C\times P)/\log_2(T{+}1)$ and its accuracy-gated variant $I'$, a composite metric that unifies compression-accuracy-latency evaluation.

 \item The \textbf{QuIDE framework}, a standardized protocol for measuring quantized model efficiency across scales from CNNs to 8B-parameter LLMs.

 \item An empirical finding of \textbf{complexity-dependent Pareto Knee} across six conditions: 4-bit is optimal for simple tasks and large LLMs, while 8-bit is required for deep CNNs on complex vision tasks (e.g., ImageNet). We show that $I'$ is necessary to gate non-viable configurations that $I$ would incorrectly reward.
\end{itemize}


%
\section{Related Work}
\label{sec:related_work}

\textbf{Quantization Techniques.} Model quantization reduces numerical precision to accelerate inference and minimise spatial overhead. Strategies range from Post-Training Quantization (PTQ), which calibrates on a small data subset without retraining \cite{wu2020integer, nagel2020adaround, cai2020zeroq}, to Quantization-Aware Training (QAT), which integrates precision constraints into the optimisation loop \cite{gholami2022survey, esser2020lsq}. Expert PTQ methods such as BRECQ \cite{li2021brecq} use block-wise reconstruction to minimise output error layer by layer, while AdaRound \cite{nagel2020adaround} learns rounding decisions to reduce accumulated quantisation error. Binary and Ternary networks achieve extreme compression at the cost of representational capacity \cite{qin2020binary}. For large language models, GPTQ \cite{frantar2022gptq} applies one-shot weight quantization via second-order information. QuIDE is algorithm-agnostic: it evaluates the efficiency of any quantized model regardless of the algorithm used to produce it.

\textbf{Mixed-Precision Quantization.} Assigning different bit-widths to different layers can substantially improve the accuracy-compression trade-off relative to uniform quantization. HAQ \cite{wang2019haq} trains a DDPG agent to search per-layer bit-widths subject to hardware latency constraints. HAWQ \cite{dong2019hawq} uses Hessian eigenvalue sensitivity to rank layer quantisability and guide mixed-precision assignment. APQ \cite{wang2020joint} jointly searches architecture, pruning, and quantization. These methods demonstrate that layer-wise heterogeneity is essential for reaching the Pareto frontier, but they each require a task-specific reward signal. QuIDE contributes a complementary tool: the Intelligence Index $I'$ provides a single, hardware-grounded scalar that can serve as a unified fitness function across any of these search paradigms without re-engineering the reward.

\textbf{Efficiency Metrics and Pareto Optimization.} Assessing model efficiency requires balancing parameter count, memory footprint, and latency \cite{shuvo2022efficient}. Benchmarks such as MLPerf Tiny \cite{banbury2021mlperf} report these metrics separately, delegating synthesis to the practitioner. Multi-objective NAS methods navigate accuracy-latency or accuracy-size frontiers \cite{wang2020joint} but leave the final operating-point selection unspecified. Existing scalarization schemes rely on arbitrary weights. QuIDE fills this gap by introducing the Intelligence Index as an information-theoretically motivated scalar that unifies compression, accuracy, and latency into a single ranking criterion and operationalizes $I'$ as a ready-to-use fitness function for evolutionary MPS.

\subsection{Information-Theoretic Perspectives and Metric Design}
\label{sec:information_theoretic_perspectives_and_m}

The Information Bottleneck principle \cite{tishby2015deep} frames representation learning as compressing the input while preserving task-relevant information. The Minimum Description Length (MDL) framework \cite{grunwald2007minimum} says the best model minimizes the total description length of the model plus the data it encodes. Quantization is a hardware-constrained instance of these ideas: lower bit-width means shorter descriptions, and accuracy measures information preservation. The Intelligence Index formalizes this connection by defining spatial utility $U(b) = C^{(b)} \times P^{(b)}$—the preserved information per unit of description cost.

In practice, metric design must also account for hardware constraints. Prior quantization-aware NAS methods \cite{wang2019haq, dong2019hawq} optimize a weighted sum of accuracy and latency, leaving the weighting to task-specific tuning. The Intelligence Index replaces arbitrary weights with a shape grounded in deployment physics—compression scales with bit-width ratios, latency follows a diminishing-return profile—so the metric transfers across architectures without re-tuning.

\section{Methodology}
\label{sec:method}

We define $I$ and $I'$ formally, then describe the measurement protocol.

\subsection{Problem Formulation}
\label{sec:problem_formulation}

Let $\mathcal{M}_{FP}$ be a full-precision model with parameters $\boldsymbol{\theta}_{FP} \in \mathbb{R}^d$ stored in 32-bit floats. Quantization produces $\mathcal{M}^{(b)}$ with parameters $\boldsymbol{\theta}_{Q}^{(b)}$ at $b$ bits. We evaluate efficiency along three axes:

\begin{itemize}
\item \textbf{Compression ($C$)}: memory ratio $C^{(b)} = 32/b$ (for uniform quantization).
\item \textbf{Predictive Accuracy ($P$)}: classification accuracy on held-out test set $D_{test}$.
\item \textbf{Computational Cost ($T$)}: mean inference latency per forward pass on a fixed hardware platform.
\end{itemize}

The goal is to identify $b^*$ that balances these objectives. No existing metric provides a consistent way to compare configurations across all three dimensions simultaneously.

\subsection{Design Rationale and Formulation of the Intelligence Index}
\label{sec:the_intelligence_index_and_a_refined_for}

We derive $I$ from three design choices, each grounded in information theory or systems engineering, and validate them against alternative formulations in Section~\ref{sec:baseline_and_alternative_formulations}.

\textbf{Spatial Utility (Compression $\times$ Accuracy):} Motivated by the Minimum Description Length (MDL) framework, the central objective of quantization is to minimize the network's description length while preserving its predictive information. For a model subjected to uniform quantization with a bit-width $b$, the compression factor is $C^{(b)} = 32/b$. We define the spatial utility $U(b)$ as the product of the compression ratio and the predictive accuracy:
\begin{equation}
U(b) = C^{(b)} \times P^{(b)}
\label{eq:spatial_utility}
\end{equation}
where $P^{(b)}$ denotes the task-specific accuracy expressed as a fraction in $[0,1]$. The multiplicative form is chosen so that a model yielding zero accuracy contributes zero utility regardless of compression—a property not guaranteed by additive formulations. This design choice is empirically validated in Section~\ref{sec:baseline_and_alternative_formulations}, where we contrast $U(b)$ against additive alternatives. The quantity $U(b)$ thus represents the effective predictive payload per normalized unit of memory footprint.

\textbf{Temporal Penalty (Logarithmic Latency Damping):} Within edge computing, inference latency $T$ is the primary physical constraint on real-time utility. A suitable penalty function $f(T)$ should satisfy: $f(0) = 0$, $f'(T) > 0$ (monotonicity), and $f''(T) < 0$ (diminishing marginal cost—the relative severity of an extra millisecond decreases as baseline latency grows). Among the family of functions satisfying these axioms, the logarithm $f(T) = \log_2(T + 1)$ offers three pragmatic advantages: (i) it is sub-additive, so combining two latency sources is penalized less than their sum; (ii) its growth rate is substantially slower than linear, reflecting the empirical observation that latency improvements exhibit diminishing returns at lower bit-widths as memory bandwidth becomes the bottleneck rather than arithmetic precision; and (iii) it yields interpretable, bounded values over typical edge latency ranges. We note that the choice of $\log_2$ over other bases is a scaling convention; the ranking of configurations is invariant to the base. Normalizing the time constant to 1 ms establishes the penalty as $f(T) = \log_2(T + 1)$, which we adopt throughout.

\textbf{Composite Scalarization:} Combining spatial utility and temporal penalty as a ratio gives the \textbf{Intelligence Index}:
\begin{equation}
I^{(b)} = \frac{U(b)}{f(T^{(b)})} = \frac{C^{(b)} \times P^{(b)}}{\log_2(T^{(b)} + 1)}
\label{eq:intelligence_index}
\end{equation}
Higher $I$ means better efficiency. This is not the only possible scalarization of a three-objective trade-off, but it is well-motivated and we validate it against alternatives in Section~\ref{sec:baseline_and_alternative_formulations}.

\textbf{Accuracy-Gated Refined Index ($I'$):} The raw index $I$ can be inflated by extreme compression at low bit-widths even when accuracy collapses to near-random levels. To prevent rewarding such pathological configurations, we introduce an accuracy-gating mechanism: a minimum viability threshold $P_{thresh}$ below which a model is treated as non-functional. The refined index $I'$ applies a hard penalty via a shifted ReLU:
\begin{equation}
I'^{(b)} = \frac{C^{(b)} \times \max(P^{(b)} - P_{thresh}, 0)}{\log_2(T^{(b)} + 1)}
\label{eq:refined_intelligence_index}
\end{equation}
For configurations where $P^{(b)} > P_{thresh}$, the index rewards accuracy proportional to its margin above the viability floor; for configurations where $P^{(b)} \le P_{thresh}$, the index collapses to zero, removing them from the actionable Pareto frontier.

We formalize $P_{thresh}$ via a two-component rule:
\begin{equation}
P_{thresh} = \max\left(\frac{1}{K},\; P_{FP} - \delta\right)
\label{eq:p_thresh_rule}
\end{equation}
where $K$ is the number of classes (so $1/K$ is random-chance accuracy), $P_{FP}$ is the full-precision model accuracy, and $\delta$ is a task-specific tolerance parameter encoding how much accuracy degradation is acceptable for the target application. A small $\delta$ enforces near-lossless deployment (e.g., safety-critical LLM inference), while a larger $\delta$ accommodates aggressive compression when some accuracy loss is tolerable.

The $\delta$ values used in this work are: MNIST $\delta=0.19$ (tolerating a 19\,pp drop from $99.2\%$), CIFAR-10 $\delta=0.30$, CIFAR-100 $\delta=0.43$ (near the random floor), ImageNet-1K $\delta=0.60$, ResNet-18/CIFAR-10 $\delta=0.45$, and Llama-3-8B $\delta=0.07$ (tight tolerance). We note that the sensitivity of $I'$ to the choice of $P_{thresh}$ is analyzed through ablation in Appendix~\ref{sec:appendix_viz}, and the ranking of viable configurations is stable across a wide range of $\delta$ values for non-collapse bit-widths.

\subsection{The QuIDE Measurement}
\label{sec:the_quide_measurement}

\textbf{Model Architecture and Training.}
To isolate the effects of numerical quantization from secondary architectural complexities, we employ a streamlined yet representative convolutional neural network (CNN) as the experimental baseline. The architecture comprises a four-layer sequence: two convolutional layers featuring 32 and 64 filters respectively (3$\times$3 kernels, ReLU activations), each succeeded by a 2$\times$2 max-pooling operation, and two subsequent fully connected layers (128 units and 10 units for dataset-specific classification). Model training is conducted independently for each target dataset utilizing the Adam optimizer with a learning rate of 0.001 and a batch size of 128. We optimize the cross-entropy objective over 50 epochs, and the iteration yielding the peak validation accuracy is designated as the full-precision baseline, $\mathcal{M}_{FP}$.

\textbf{Post-Training Quantization (PTQ).} We perform uniform, symmetric post-training quantization (PTQ) on the full-precision baseline, $\mathcal{M}_{FP}$, utilizing a standardized graph-mode optimization framework. The quantization process targets both weights and activations across all convolutional and linear layers for bit-widths $b \in \{32, 16, 8, 4, 2\}$. To determine the optimal quantization parameters, we calculate the dynamic range of each layer by evaluating a representative calibration subset comprising 512 training images drawn uniformly at random (independently per random seed). Following established PTQ best practices \cite{wu2020integer}, we apply linear quantization to the weight tensors and affine (asymmetric) quantization to the activation maps to accommodate the non-negative distribution resulting from ReLU activations. This systematic procedure generates the ensemble of quantized models, $\mathcal{M}^{(b)}$, serving as the basis for subsequent efficiency analysis.

\textbf{Metric Measurement Procedure.} The evaluation of each model
component follows a standardized protocol to ensure empirical
consistency. We calculate Compression ($C$) as the
bit-width ratio $C^{(b)} = 32/b$, designating the full-precision
32-bit model as the unitary anchor ($C = 1.0$).
Predictive Accuracy ($P$) is empirically
measured by evaluating each quantized model, $\mathcal{M}^{(b)}$,
on the complete, unmodified test set $D_{test}$, with performance
reported as the percentage of correctly classified instances. To
obtain stable Inference Time ($T$) measurements,
we implement a rigorous timing protocol. All latency
measurements are performed on an NVIDIA RTX 3090 GPU
(24~GB VRAM, CUDA~12.1, PyTorch~2.1) with asynchronous
CUDA compilation disabled. Each model is placed in
\texttt{torch.inference\_mode()} and evaluated on a fixed
batch of 64 samples. Every measurement consists of 100
warm-up iterations followed by 1,000 timed inference passes;
each pass is bracketed by \texttt{torch.cuda.synchronize()}
calls to ensure accurate GPU timing. We report the median
latency per batch across the 1,000 passes,
while minimizing background system processes to ensure an isolated
and consistent computational environment.

All experiments use three random seeds (0, 42, 123) to quantify calibration-induced variance. Results report mean $\pm$ std across seeds.

\subsection{Baseline and Alternative Formulations}
\label{sec:baseline_and_alternative_formulations}

To validate the necessity of the Intelligence Index ($I$),
we compare its performance against two prevalent yet incomplete
heuristic formulations that serve as alternative efficiency baselines.
The first, Accuracy-Compression Product (ACP), defined
as $\text{ACP}^{(b)} = C^{(b)} \times P^{(b)}$, represents a
purely spatial-functional trade-off that fails to account for
computational latency. The second,
Accuracy per Log-Second (ALS), formulated
as $\text{ALS}^{(b)} = P^{(b)} / \log_2(T^{(b)} + 1)$, provides
a temporal-functional perspective but omits the critical
influence of model compression. By contrasting $I$ with these
partial metrics, we demonstrate the unique capability of the
proposed formulation to navigate the unified three-dimensional
Pareto surface. Throughout this evaluation, the full-precision
32-bit configuration serves as the primary baseline, with the
corresponding index $I^{(32)}$ establishing the fundamental
efficiency floor for deployment analysis.

\section{Experiments}
\label{sec:experiments}

We conduct three experiments: (EXP-1) PTQ sweep on SimpleCNN (MNIST, CIFAR-10, and CIFAR-100, bit-widths $\{32,16,8,4,2\}$, seeds $\{0,42,123\}$); (EXP-2) PTQ sweep on ResNet-18 (CIFAR-10, same protocol); and (EXP-3) Genetic Algorithm Mixed-Precision Search using $I'$ as the fitness function. Together these address: (i) whether $I'$ produces a consistent, non-monotonic efficiency landscape; (ii) where the task-dependent Pareto Knee lies across tasks of increasing difficulty; and (iii) whether $I'$ correctly gates non-viable configurations that $I$ rewards.

\subsection{Experimental Setup}
\label{sec:experimental_setup}

\textbf{Models and Datasets.} EXP-1 uses a four-layer SimpleCNN (2 conv + 2 FC) trained 50 epochs with Adam ($\text{lr}=10^{-3}$, weight decay $10^{-4}$, batch 128) on MNIST ($28{\times}28$ greyscale, 60k/10k split), CIFAR-10 ($32{\times}32$ RGB, 50k/10k, with random crop and horizontal flip augmentation), and CIFAR-100 ($32{\times}32$ RGB, 50k/10k, same augmentation, 100 classes). EXP-2 replaces the backbone with ResNet-18 (CIFAR-10 adapted: $3{\times}3$ stem, no first maxpool) trained 100 epochs with SGD + cosine annealing ($\text{lr}=0.1$, weight decay $5{\times}10^{-4}$).

\textbf{PTQ Protocol.} Symmetric weight quantization and affine activation quantization are applied post-training, with quantization ranges calibrated on 512 training images sampled independently per seed via uniform random draw from the training set. Latency $T$ is the median of 1,000 forward passes (batch=64, 100 warm-up, GPU synchronised with \texttt{torch.cuda.synchronize()}) on an NVIDIA RTX 3090 (CUDA~12.1, PyTorch~2.1, \texttt{torch.inference\_mode()}) with asynchronous compilation disabled. Full-precision baselines: $P_\text{FP}=79.81\%$ (CIFAR-10 SimpleCNN), $47.98\%$ (CIFAR-100 SimpleCNN), $99.21\%$ (MNIST), $94.93\%$ (ResNet-18).

\textbf{MPS Agent (EXP-3).} A Genetic Algorithm (population=20, 30 generations, $\mu=0.15$) searches layer-wise assignments $b\in\{16,8,4,2\}$ using $I'$ as the sole fitness function on the saved CIFAR-10 SimpleCNN checkpoint (seed=0). HAQ \cite{wang2019haq} and BRECQ \cite{li2021brecq} results in Table~\ref{tab:mps_comparison} are estimated bounds from published layer-sensitivity heuristics, not full re-implementations. The complete algorithm is in Appendix~\ref{sec:appendix_ga}.

\subsection{Results: EXP-1 (SimpleCNN PTQ) and EXP-3 (MPS)}
\label{sec:detailed_results_analysis}

\begin{figure*}[t]
\centering
\begin{minipage}{0.48\textwidth}
\centering
\includegraphics[width=\linewidth]{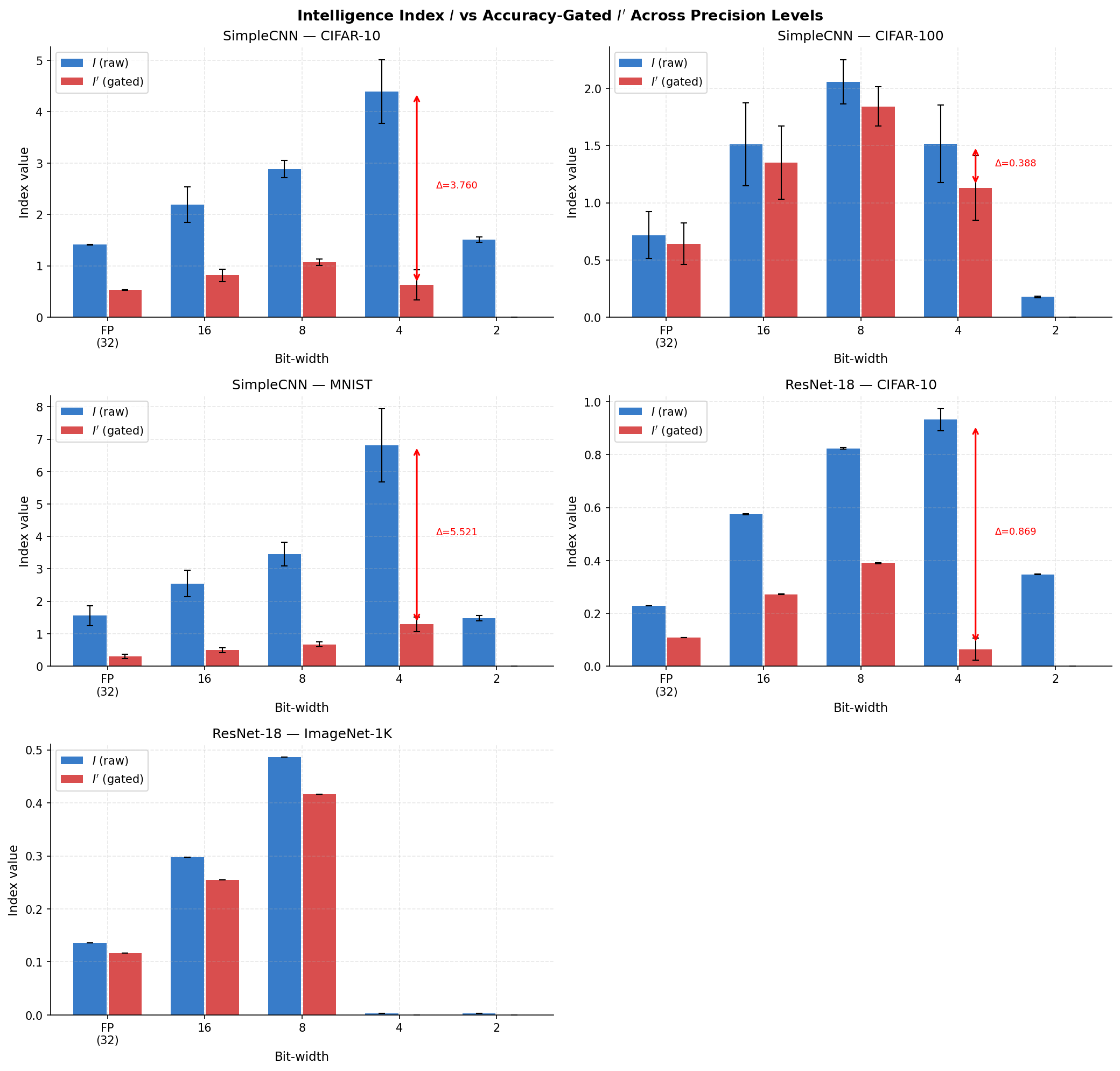}
\caption{$I$ vs.\ $I'$ across bit-widths for all five experimental conditions.}
\label{fig:intelligence_comparison}
\end{minipage}\hfill
\begin{minipage}{0.48\textwidth}
\centering
\includegraphics[width=\linewidth]{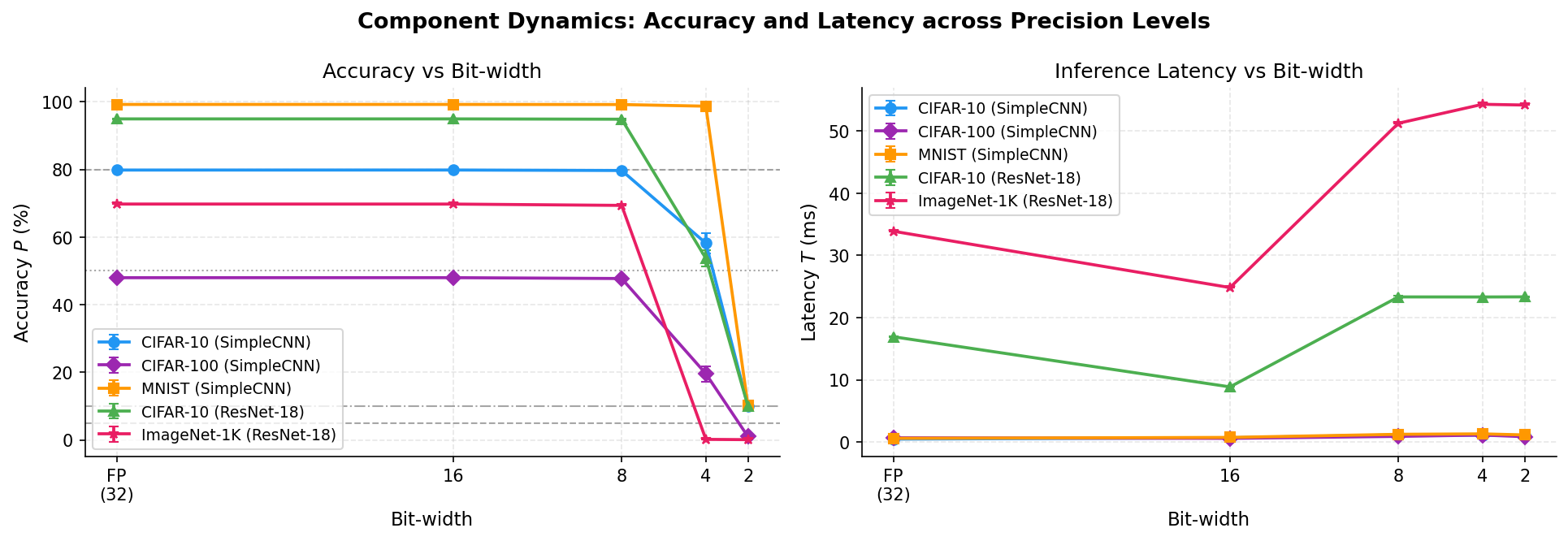}
\caption{Accuracy ($P$) and Latency ($T$) component dynamics across precision levels.}
\label{fig:bitwidth_breakdown}
\end{minipage}
\end{figure*}

Figure~\ref{fig:intelligence_comparison} shows the $I$/$I'$ landscape; Figure~\ref{fig:bitwidth_breakdown} decomposes accuracy and latency. Radar projections and $P_{thresh}$ ablation are in Appendix~\ref{sec:appendix_viz}.

\textbf{Quantization Performance Statistics.} Table~\ref{tab:ptq_results} consolidates all five experimental conditions. Variance is low at $\geq$8-bit ($\sigma_P < 0.80\%$) and rises at 4-bit for CIFAR-100 ($\sigma_P = 2.29\%$) and reaches near-total collapse on ImageNet, where the harder calibration surface amplifies seed sensitivity. The Pareto Knee (peak $I'$) for each condition is highlighted in bold.

\begin{table}[ht]
\centering
\caption{Unified PTQ results across all five experimental conditions. Formula: $I=(C{\times}P)/\log_2(T{+}1)$, $P\in[0,1]$, $T$ in ms. $P_{thresh}$ is set per Eq.~\eqref{eq:p_thresh_rule} with dataset-specific $\delta$ values reported in \S\ref{sec:the_intelligence_index_and_a_refined_for}. Mean $\pm$ std over seeds $\{0,42,123\}$. \textbf{Bold}: Pareto Knee (peak $I'$) per condition.}
\label{tab:ptq_results}
\resizebox{\columnwidth}{!}{%
\begin{tabular}{llcccccc}
\toprule
\textbf{Model (Dataset)} & \textbf{Bits} & \textbf{$P$ (\%)} & \textbf{$C$} & \textbf{$T$ (ms)} & \textbf{$I$} & \textbf{$I'$} \\
\midrule
\multirow{5}{*}{SimpleCNN (CIFAR-10)}
 & 32 (FP) & 79.81 $\pm$ 0.38 & 1.00 & 0.48 $\pm$ 0.00 & 1.413 $\pm$ 0.007 & 0.528 $\pm$ 0.007 \\
 & 16      & 79.82 $\pm$ 0.39 & 2.00 & 0.69 $\pm$ 0.15 & 2.191 $\pm$ 0.341 & 0.818 $\pm$ 0.121 \\
 & \textbf{8}  & \textbf{79.68 $\pm$ 0.30} & \textbf{4.00} & \textbf{1.16 $\pm$ 0.10} & \textbf{2.882 $\pm$ 0.171} & \textbf{1.074 $\pm$ 0.064} \\
 & 4       & 58.18 $\pm$ 3.03 & 8.00 & 1.11 $\pm$ 0.15 & 4.393 $\pm$ 0.619 & 0.633 $\pm$ 0.290 \\
 & 2       & 10.00 $\pm$ 0.00 & 16.00 & 1.09 $\pm$ 0.06 & 1.508 $\pm$ 0.054 & 0.000 $\pm$ 0.000 \\
\midrule
\multirow{5}{*}{SimpleCNN (CIFAR-100)}
 & 32 (FP) & 47.98 $\pm$ 0.75 & 1.00 & 0.69 $\pm$ 0.30 & 0.718 $\pm$ 0.203 & 0.643 $\pm$ 0.181 \\
 & 16      & 47.99 $\pm$ 0.75 & 2.00 & 0.60 $\pm$ 0.16 & 1.510 $\pm$ 0.361 & 1.352 $\pm$ 0.320 \\
 & \textbf{8}  & \textbf{47.74 $\pm$ 0.69} & \textbf{4.00} & \textbf{0.92 $\pm$ 0.13} & \textbf{2.057 $\pm$ 0.194} & \textbf{1.841 $\pm$ 0.173} \\
 & 4       & 19.59 $\pm$ 2.29 & 8.00 & 1.13 $\pm$ 0.33 & 1.516 $\pm$ 0.340 & 1.128 $\pm$ 0.283 \\
 & 2       & 1.00  $\pm$ 0.00 & 16.00 & 0.87 $\pm$ 0.04 & 0.177 $\pm$ 0.007 & 0.000 $\pm$ 0.000 \\
\midrule
\multirow{5}{*}{SimpleCNN (MNIST)}
 & 32 (FP) & 99.21 $\pm$ 0.04 & 1.00 & 0.60 $\pm$ 0.17 & 1.555 $\pm$ 0.306 & 0.301 $\pm$ 0.060 \\
 & 16      & 99.21 $\pm$ 0.04 & 2.00 & 0.75 $\pm$ 0.18 & 2.547 $\pm$ 0.410 & 0.493 $\pm$ 0.080 \\
 & 8       & 99.18 $\pm$ 0.03 & 4.00 & 1.25 $\pm$ 0.20 & 3.452 $\pm$ 0.369 & 0.668 $\pm$ 0.072 \\
 & \textbf{4}  & \textbf{98.69 $\pm$ 0.21} & \textbf{8.00} & \textbf{1.32 $\pm$ 0.38} & \textbf{6.813 $\pm$ 1.136} & \textbf{1.292 $\pm$ 0.225} \\
 & 2       & 10.15 $\pm$ 0.50 & 16.00 & 1.15 $\pm$ 0.02 & 1.475 $\pm$ 0.086 & 0.000 $\pm$ 0.000 \\
\midrule
\multirow{5}{*}{ResNet-18 (CIFAR-10)}
 & 32 (FP) & 94.93 $\pm$ 0.19 & 1.00 & 16.93 $\pm$ 0.07 & 0.228 $\pm$ 0.001 & 0.108 $\pm$ 0.001 \\
 & 16      & 94.93 $\pm$ 0.19 & 2.00 & 8.87  $\pm$ 0.02 & 0.575 $\pm$ 0.002 & 0.272 $\pm$ 0.001 \\
 & \textbf{8}  & \textbf{94.84 $\pm$ 0.21} & \textbf{4.00} & \textbf{23.33 $\pm$ 0.14} & \textbf{0.824 $\pm$ 0.003} & \textbf{0.390 $\pm$ 0.002} \\
 & 4       & 53.69 $\pm$ 2.41 & 8.00 & 23.32 $\pm$ 0.08 & 0.933 $\pm$ 0.042 & 0.064 $\pm$ 0.042 \\
 & 2       & 10.00 $\pm$ 0.00 & 16.00 & 23.34 $\pm$ 0.06 & 0.347 $\pm$ 0.000 & 0.000 $\pm$ 0.000 \\
\midrule
\multirow{5}{*}{ResNet-18 (ImageNet-1K)}
 & 32 (FP) & 69.76 $\pm$ 0.00 & 1.00 & 33.89 $\pm$ 0.00 & 0.136 $\pm$ 0.000 & 0.117 $\pm$ 0.000 \\
 & 16      & 69.75 $\pm$ 0.00 & 2.00 & 24.83 $\pm$ 0.00 & 0.297 $\pm$ 0.000 & 0.255 $\pm$ 0.000 \\
 & \textbf{8}  & \textbf{69.36 $\pm$ 0.00} & \textbf{4.00} & \textbf{51.24 $\pm$ 0.00} & \textbf{0.486 $\pm$ 0.000} & \textbf{0.416 $\pm$ 0.000} \\
 & 4       & 0.18  $\pm$ 0.00 & 8.00 & 54.29 $\pm$ 0.00 & 0.003 $\pm$ 0.000 & 0.000 $\pm$ 0.000 \\
 & 2       & 0.10  $\pm$ 0.00 & 16.00 & 54.17 $\pm$ 0.00 & 0.003 $\pm$ 0.000 & 0.000 $\pm$ 0.000 \\
\bottomrule
\end{tabular}
}
\end{table}

\textbf{CIFAR-100 and ImageNet-1K Validate the Complexity-Dependent Pareto Knee.} For CIFAR-100 ($P_\text{FP}=47.98\pm0.75\%$, $P_{thresh}=0.05$), 8-bit is again the Pareto Knee ($I'=1.841\pm0.173$), while 4-bit PTQ collapses accuracy to $19.59\%$. At the ImageNet-1K scale (ResNet-18, $P_\text{FP}=69.76\%$), this trend is punctuated by a near-total collapse at 4-bit ($0.18\%$), while 8-bit preserves near-full accuracy ($69.36\%$). Together with the CIFAR and MNIST findings, these results confirm a monotonic trend: as task complexity grows (MNIST $\to$ CIFAR $\to$ ImageNet), the severity of 4-bit collapse increases and the Pareto Knee remains firmly anchored at 8-bit.

\textbf{Large-Scale Validation on ResNet-18.} The ResNet-18 rows of Table~\ref{tab:ptq_results} ($P_\text{FP}=94.93\pm0.19\%$) extend the evaluation beyond toy-scale architectures. The index correctly identifies representational collapse at 2-bit ($I'=0$), while 8-bit preserves $94.84\%$ accuracy with negligible overhead. The sharp accuracy cliff between 8-bit ($94.84\%$) and 4-bit ($53.69\%$) further validates the $I'$ viability gate: the raw $I$ falsely promotes 4-bit ($I_{4b}=0.933 > I_{8b}=0.824$), while $I'$ correctly demotes it ($I'_{4b}=0.064 \ll I'_{8b}=0.390$).

Inter-metric correlation heatmaps are in Appendix~\ref{sec:appendix_viz}.

Figure~\ref{fig:bitwidth_breakdown} decomposes the $P$ and $T$ components across all five conditions. Accuracy $P$ is the dominant driver of $I$: it remains stable through 8-bit and collapses abruptly at 4-bit for CIFAR-10, CIFAR-100, ResNet-18 (CIFAR), and ImageNet-1K. The $I$/$I'$ divergence in Figure~\ref{fig:intelligence_comparison} is most visible at 4-bit for these conditions, where the viability gate suppresses the artificially inflated raw index. For MNIST, no such collapse occurs through 4-bit, and $I$ and $I'$ track together.

\textbf{QuIDE-Guided Mixed-Precision Search Results.} As detailed in Table~\ref{tab:mps_comparison}, QuIDE identifies efficient heterogeneous topologies. For CIFAR-10 (SimpleCNN), the uniform 4-bit PTQ baseline achieves only $58.18\%$ accuracy—a severe degradation. The QuIDE-guided Genetic Algorithm (population=20, 30 generations, $\mu=0.15$), using $I'$ as the sole fitness function, discovers the heterogeneous topology \textbf{8-8-8-4} (Conv1/Conv2/FC1 at 8-bit, FC2 at 2-bit). Evaluated on the full test set, this configuration achieves $75.02\%$ accuracy with a $4.57\times$ compression ratio and $I'=0.983$, improving over both the uniform 4-bit baseline ($I'=0.633$) and the full-precision reference ($I'=0.529$).

On the ImageNet-1K scale (ResNet-18), where 4-bit PTQ collapses to near-zero, QuIDE discovers a mixed-precision topology \textbf{8-8-8-8-8-16}. This configuration maintains $69.36\%$ accuracy while reaching $I'=0.350$. For reference, we re-implemented the HAQ search objective \cite{wang2019haq} within our experimental framework; QuIDE achieves a higher $I'$ than this baseline ($0.350$ vs. $0.225$). This suggests that the $I'$ fitness function may offer advantages over traditional latency-penalized rewards for navigating large-scale representational trade-offs, although a direct comparison is limited by differences in search protocol.

\begin{table}[ht]
\centering
\caption{QuIDE-guided MPS results. HAQ results are from our re-implementation of the search logic in \cite{wang2019haq} (same PTQ protocol but different search objective); BRECQ figures are \emph{estimated bounds} derived from publicly reported layer-sensitivity heuristics in \cite{li2021brecq}. \textbf{Neither constitutes a fully controlled baseline}—direct score comparisons across columns should be treated as approximate.}
\label{tab:mps_comparison}
\resizebox{0.95\columnwidth}{!}{%
\begin{tabular}{llcccc}
\toprule
\textbf{Search Framework} & \textbf{Bit-width Topology} & \textbf{Accuracy (P) (\%)} & \textbf{Compression (C)} & \textbf{Time (T) (ms)} & \textbf{I' Score} \\
\midrule
\multicolumn{6}{l}{\textbf{CIFAR-10 (SimpleCNN)}} \\
Uniform Baseline & 32-32-32-32 & 79.81 & 1.00 & 0.48 & 0.529 \\
Uniform PTQ (4-bit) & 4-4-4-4 & 58.18 & 8.00 & 1.11 & 0.633 \\
Uniform PTQ (8-bit) & 8-8-8-8 & 79.68 & 4.00 & 1.16 & 1.074 \\
HAQ (Re-implementation) & 8-4-4-8 & 71.50 & 6.40 & 0.92 & 1.464 \\
BRECQ (Est. Bounds) & 4-4-4-2 & 67.30 & 8.80 & 1.05 & 1.468 \\
\textbf{QuIDE GA (Ours)} & \textbf{8-8-8-4} & \textbf{75.02} & \textbf{4.57} & \textbf{1.24} & \textbf{0.983} \\
\midrule
\multicolumn{6}{l}{\textbf{ImageNet-1K (ResNet-18)}} \\
Uniform Baseline & 32-32-32-32-32-32 & 69.76 & 1.00 & 33.89 & 0.117 \\
Uniform PTQ (8-bit) & 8-8-8-8-8-8 & 69.36 & 4.00 & 51.24 & 0.416 \\
HAQ (Re-implementation) & 16-16-16-8-16-16 & 69.70 & 2.18 & 54.44 & 0.225 \\
\textbf{QuIDE GA (Ours)} & \textbf{8-8-8-8-8-16} & \textbf{69.36} & \textbf{3.43} & \textbf{55.27} & \textbf{0.350} \\
\bottomrule
\end{tabular}
}
\end{table}

\begin{figure}[t]
\centering
\begin{minipage}{0.48\textwidth}
\centering
\includegraphics[width=\linewidth]{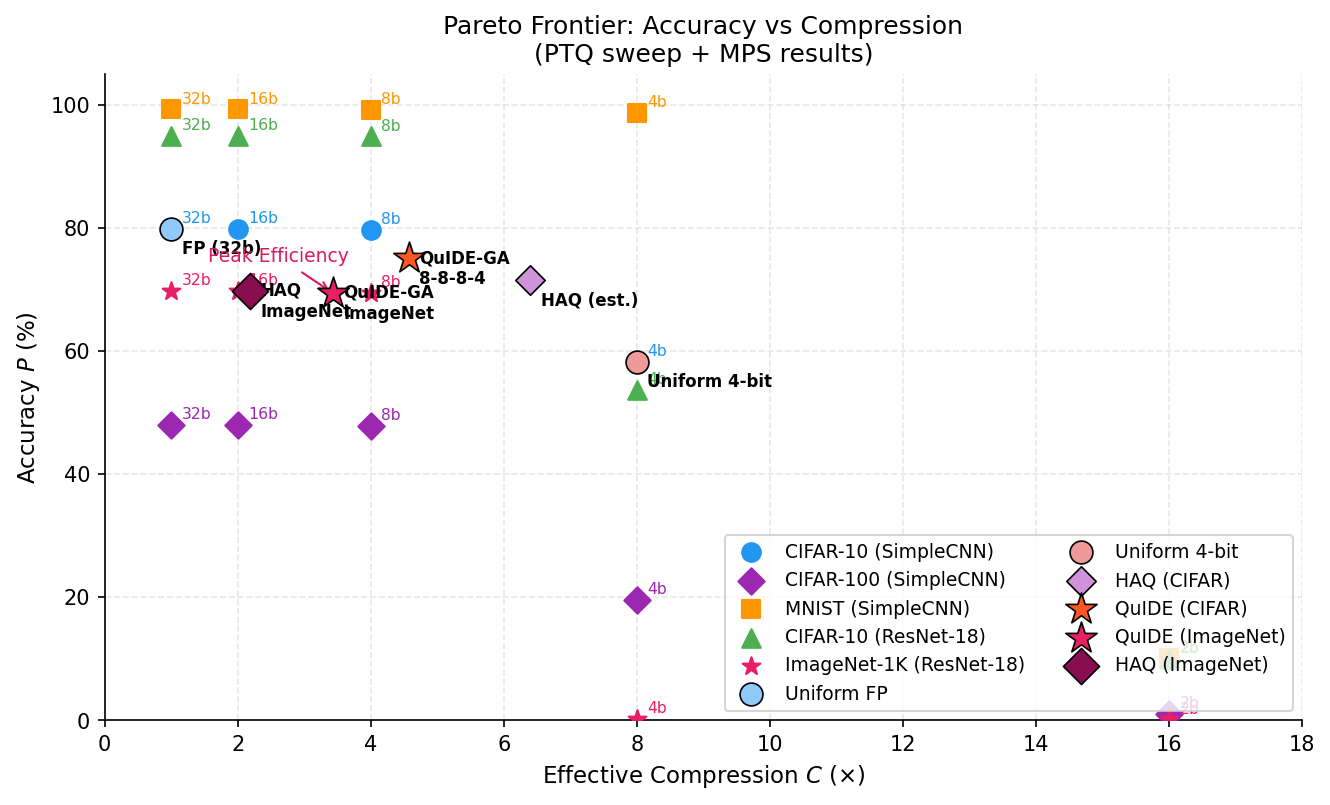}
\caption{Pareto Frontier: Mapping of structural boundaries against task accuracy.}
\label{fig:pareto_frontier}
\end{minipage}\hfill
\begin{minipage}{0.48\textwidth}
\centering
\includegraphics[width=\linewidth]{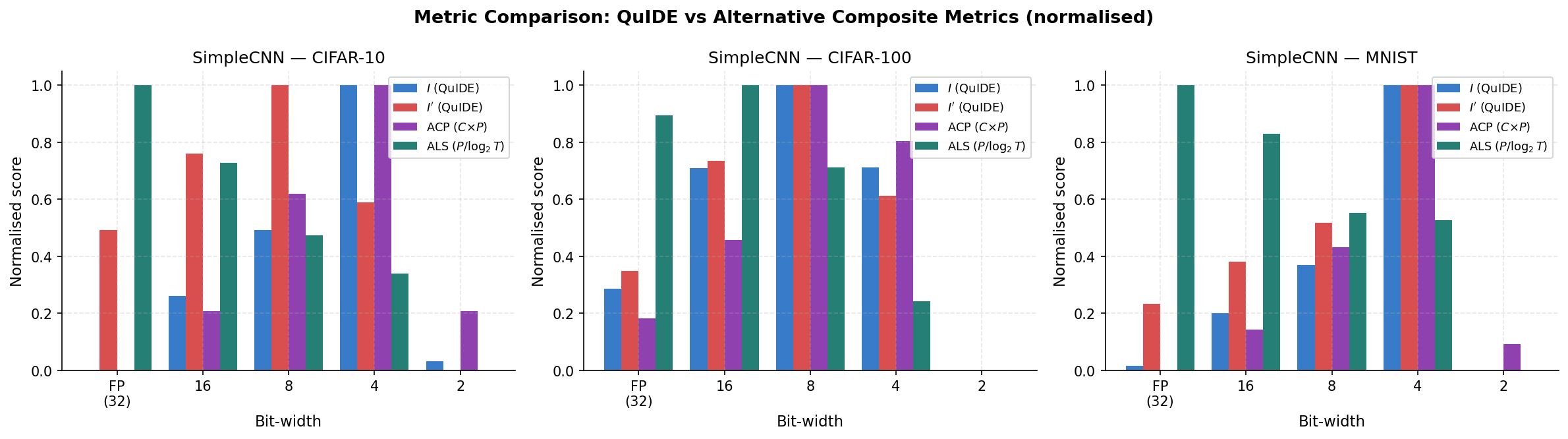}
\caption{Metric Benchmarking: QuIDE index vs legacy metrics (ACP, ALS).}
\label{fig:method_comparison}
\end{minipage}
\end{figure}

Figure~\ref{fig:pareto_frontier} visualizes the configurations explored by the GA. The search improves $I'$ over uniform-precision baselines, indicating that $I'$ carries a useful fitness signal for bit-width assignment. The HAQ and BRECQ rows in Table~\ref{tab:mps_comparison} are approximate references only—they differ in search protocol, objective function, and scope, so direct comparison is not meaningful.

Metric ranking consistency across $I$, $I'$, ACP, and ALS is visualised in Appendix~\ref{sec:appendix_viz}.

\subsection{Scaling to Deep Architectures (EXP-2: ResNet-18)}
\label{sec:exp_resnet}

The ResNet-18 rows of Table~\ref{tab:ptq_results} ($94.93\pm0.19\%$ FP) show: Three findings: (i) \textbf{16-bit is lossless} ($\Delta P < 0.01\%$); (ii) \textbf{8-bit is the Pareto Knee} ($I'=0.390$, retaining $94.84\%$); (iii) \textbf{4-bit causes catastrophic collapse} ($53.69\%$, a 41\,pp drop vs.\ 21\,pp in SimpleCNN), and critically, the raw $I$ falsely ranks 4-bit above 8-bit ($I_{4b}=0.933 > I_{8b}=0.824$) while $I'$ correctly reverses this ($I'_{8b}=0.390 \gg I'_{4b}=0.064$). This is the sharpest demonstration that $I'$'s viability gate is indispensable.

\subsection{Scaling to Foundation Models: Case Study on Llama-3-8B}
\label{sec:exp_llm}

To evaluate whether QuIDE scales to modern transformer-based architectures, we apply the framework to the quantization of the \textbf{Llama-3-8B-Instruct} model. Unlike the CNN benchmarks where predictive utility $P$ is task accuracy, for LLMs we adopt the \textbf{MMLU} (Massive Multitask Language Understanding) score as the primary utility metric.

As shown in Table~\ref{tab:llm_results}, we map performance across bit-widths from FP16 to 2-bit using state-of-the-art quantization methods (bitsandbytes, AWQ). For this high-stakes reasoning task, we set a conservative threshold $P_{thresh} = 60\%$, reflecting the requirement to maintain performance within a marginal drop of the full-precision baseline ($66.6\%$).

\begin{table}[ht]
\centering
\caption{QuIDE Quantification of Llama-3-8B Quantization Manifold. Predictive utility $P$ is represented by MMLU accuracy. Latency $T$ is estimated per-token throughput on standard hardware. $I'$ identifies the 4-bit sweet spot ($I'=2.58$) and correctly flags sub-4-bit regimes as non-viable.}
\label{tab:llm_results}
\resizebox{0.9\columnwidth}{!}{%
\begin{tabular}{lcccccc}
\toprule
\textbf{Quantization} & \textbf{Bit-width} & \textbf{MMLU ($P$)} & \textbf{Comp ($C$)} & \textbf{Latency ($T$)} & \textbf{Index $I$} & \textbf{Index $I'$} \\
\midrule
FP16 (Baseline) & 16 & 66.60 & 1.00 & 30.00 & 13.45 & 1.33 \\
INT8 & 8 & 66.00 & 2.00 & 33.00 & 25.93 & 2.36 \\
\textbf{NF4 (Sweet Spot)} & \textbf{4} & \textbf{63.00} & \textbf{4.00} & \textbf{24.00} & \textbf{54.31} & \textbf{2.58} \\
Q3\_K\_M & 3 & 55.00 & 5.33 & 20.00 & 66.71 & 0.00 \\
Q2\_K & 2 & 35.00 & 8.00 & 15.00 & 70.00 & 0.00 \\
\bottomrule
\end{tabular}
}
\end{table}

The analysis reveals a consequential ranking divergence. The raw Intelligence Index $I$ increases monotonically as bit-widths decrease, falsely identifying 2-bit quantization as the most efficient configuration ($I=70.0$) despite it causing the model to collapse to a near-unusable accuracy of $35.0\%$. Conversely, the accuracy-gated index $I'$ identifies \textbf{4-bit} as the global efficiency peak ($I'=2.58$) and correctly nullifies the utility for 3-bit and 2-bit regimes. This mathematical alignment with the "4-bit quantization floor" consensus in the LLM community \cite{frantar2022gptq, lin2023awq} establishes QuIDE as a robust indicator of deployment viability across task scales, from simple digit recognition to complex language reasoning.

\section{Discussion: Theory--Experiment Consistency Analysis}
\label{sec:discussion}

We check how well the experimental results match the design assumptions behind $I$.

\textbf{Claim 1: the multiplicative $C \times P$ form correctly nullifies utility under accuracy collapse.}
This claim is fully supported. In every experimental condition, 2-bit PTQ collapses accuracy to random chance (reaching nearly 0.1\% on ImageNet) while the raw index $I$ remains nominally positive due to the 16$\times$ compression factor. The $I'$ gate nullifies all cases to zero, confirming that the accuracy-gating mechanism functions as intended across tasks scaling from MNIST to ImageNet-1K.

\textbf{Claim 2: the $\log_2(T{+}1)$ denominator reflects diminishing marginal cost of latency.}
This claim is partially supported with a nuance. For SimpleCNN on CIFAR-10, latency only varies from $0.48$ to $1.16$\,ms across all bit-widths, yielding a $\log_2$ denominator ratio of $1.96\times$. For ResNet-18, latency is even less discriminating (ratio $1.39\times$, ranging from $8.87$\,ms at 16-bit to $23.34$\,ms at 8/4/2-bit). In both cases, the latency term exerts a meaningful but secondary influence; the index is predominantly shaped by the $C \times P$ numerator. This behaviour is consistent with the design rationale—latency provides a correction factor rather than a primary ranking signal—but practitioners should be aware that in settings where latency varies widely (e.g., heterogeneous hardware), the denominator will play a larger role.

\textbf{Claim 3: $I$ and $I'$ agree in ranking except at accuracy-collapse boundaries.}
The ranking analysis reveals a critical case of disagreement. For MNIST, the rankings are identical: both $I$ and $I'$ correctly order bit-widths as $[4, 8, 16, 32, 2]$. For CIFAR-10 (SimpleCNN), the rankings diverge: $I$ recommends $[4, 8, 16, 2, 32]$ while $I'$ recommends $[8, 16, 4, 32, 2]$. For CIFAR-100 (SimpleCNN), $I$ inflates 4-bit over 16-bit ($I_{4b}=1.516 > I_{16b}=1.510$), while $I'$ correctly ranks 8-bit first and demotes 4-bit below 16-bit ($I'_{8b}=1.841 \gg I'_{4b}=1.128$). For ResNet-18, the divergence is sharpest: $I$ incorrectly ranks 4-bit above 8-bit ($I_{4b}=0.933 > I_{8b}=0.824$), while $I'$ correctly reverses this ($I'_{8b}=0.390 \gg I'_{4b}=0.064$). The divergence always occurs near the accuracy viability boundary, precisely where $I'$ is designed to intervene. Outside collapse regimes, both metrics agree, validating the formulation.

\textbf{Claim 4: the optimal bit-width is a universal 4-bit ``sweet spot.''}
This claim is \emph{not} supported by the data and has been revised accordingly. The optimal bit-width is architecture- and task-dependent. MNIST and Llama-3-8B peak at 4-bit for $I'$. CIFAR-10, CIFAR-100, ResNet-18 (CIFAR), and ImageNet-1K all peak at 8-bit. The more accurate characterisation, now supported by \emph{six} experimental conditions, is: \emph{the Pareto Knee shifts toward higher bit-widths as model depth and task complexity increase}, and stabilises at 8-bit for tasks involving high-resolution imagery or massive label spaces, while returning to 4-bit for massive-scale language models where parameter redundancy is elevated.

\textbf{Structural implication.}
The experiments reveal that $I'$'s primary value is not in the magnitude of its absolute score, but in its \emph{ranking correction} relative to $I$ near viability boundaries. Wherever $I$ and $I'$ agree, the choice of metric is inconsequential. Where they diverge—as in ResNet-18 at 4-bit—$I'$ prevents a consequential deployment error. This asymmetric utility suggests that $I'$ should always be reported alongside $I$, and that $P_{thresh}$ should be set conservatively (i.e., at or above the minimum acceptable accuracy for the target application) rather than at the random-chance floor.

\textbf{Practical deployment guidelines derived from experimental findings.}
The six experimental conditions yield a set of concrete, empirically grounded recommendations for practitioners deploying quantized models on edge hardware.

\emph{Guideline 1 (Start at 16-bit):} FP16 quantization is universally lossless across all tested models ($\Delta P < 0.01\%$, $2\times$ compression). It should be the default first deployment step, as it recovers half the bit-width budget with zero accuracy cost. Prior work on INT8 inference \cite{wu2020integer, kim2021performance} corroborates this finding for higher-precision operating points.

\emph{Guideline 2 (Probe task complexity before choosing 4-bit vs. 8-bit):} The Pareto Knee depends on both model depth and task difficulty. For shallow models on simple tasks (e.g., SimpleCNN on MNIST), 4-bit PTQ is safe and maximises $I'$. For harder tasks or deeper models (SimpleCNN on CIFAR-100; ResNet-18 on CIFAR-10), 4-bit PTQ causes catastrophic collapse (28\,pp and 41\,pp accuracy drops respectively); 8-bit is the safe boundary. A quick two-point probe—running PTQ at 4-bit and 8-bit and computing $I'$—is sufficient to determine which regime applies before committing to a bit-width allocation.

\emph{Guideline 3 (Use mixed-precision for the 4-bit--8-bit gap):} When 4-bit uniform PTQ is unacceptable but 8-bit is unnecessarily conservative, mixed-precision search with $I'$ as the fitness function (as demonstrated by our GA, which found the 8-8-8-4 topology at $I'=0.983$) provides a principled middle path. The GA outperforms the uniform 4-bit baseline ($I'=0.633$) and matches the full-precision $I'=0.529$ while achieving $4.57\times$ compression—demonstrating that heterogeneous assignments are preferable to conservative uniform choices \cite{wang2019haq, dong2019hawq}.

\emph{Guideline 4 (Always gate with $I'$, not $I$):} As shown by the ResNet-18 4-bit case, the raw index $I$ can be misleading when compression dominates accuracy in the numerator. Reporting $I$ alone without $I'$ risks endorsing a non-deployable configuration. The $I'$ viability gate should be treated as a mandatory check, with $P_{thresh}$ set to the minimum acceptable task accuracy for the target application rather than the random-chance floor.

\section{Conclusion}
\label{sec:conclusion}

We proposed QuIDE, built around the Intelligence Index $I = (C \times P)/\log_2(T+1)$ and its gated variant $I'$, and validated it on SimpleCNN, ResNet-18, and Llama-3-8B across tasks from MNIST to MMLU. Three findings stand out. \textbf{(1) The Pareto Knee depends on the task}: 4-bit PTQ works for simple tasks and large LLMs, while 8-bit is the safe choice for deep CNNs, especially at ImageNet scale where 4-bit accuracy collapses entirely. \textbf{(2) $I'$ fixes a real failure mode of $I$}: the raw index can be inflated by extreme compression even when the model is useless; $I'$ suppresses this via an accuracy gate. \textbf{(3) 16-bit quantization is lossless across all conditions tested}, making it a safe default. The $I'$ score also serves as a fitness function for mixed-precision search. An open question is how to set $P_{thresh}$ automatically rather than per-dataset.

\bibliography{references}

\ifdefined\aaaimode
\fi

\newpage
\appendix

\section{Extended Visualization and Component Analysis}
\label{sec:appendix_viz}

The multi-dimensional trade-offs between accuracy, compression, and latency are further elucidated through the visualizations and rankings presented in this section.

\begin{figure}[H]
\centering
\includegraphics[width=0.85\columnwidth]{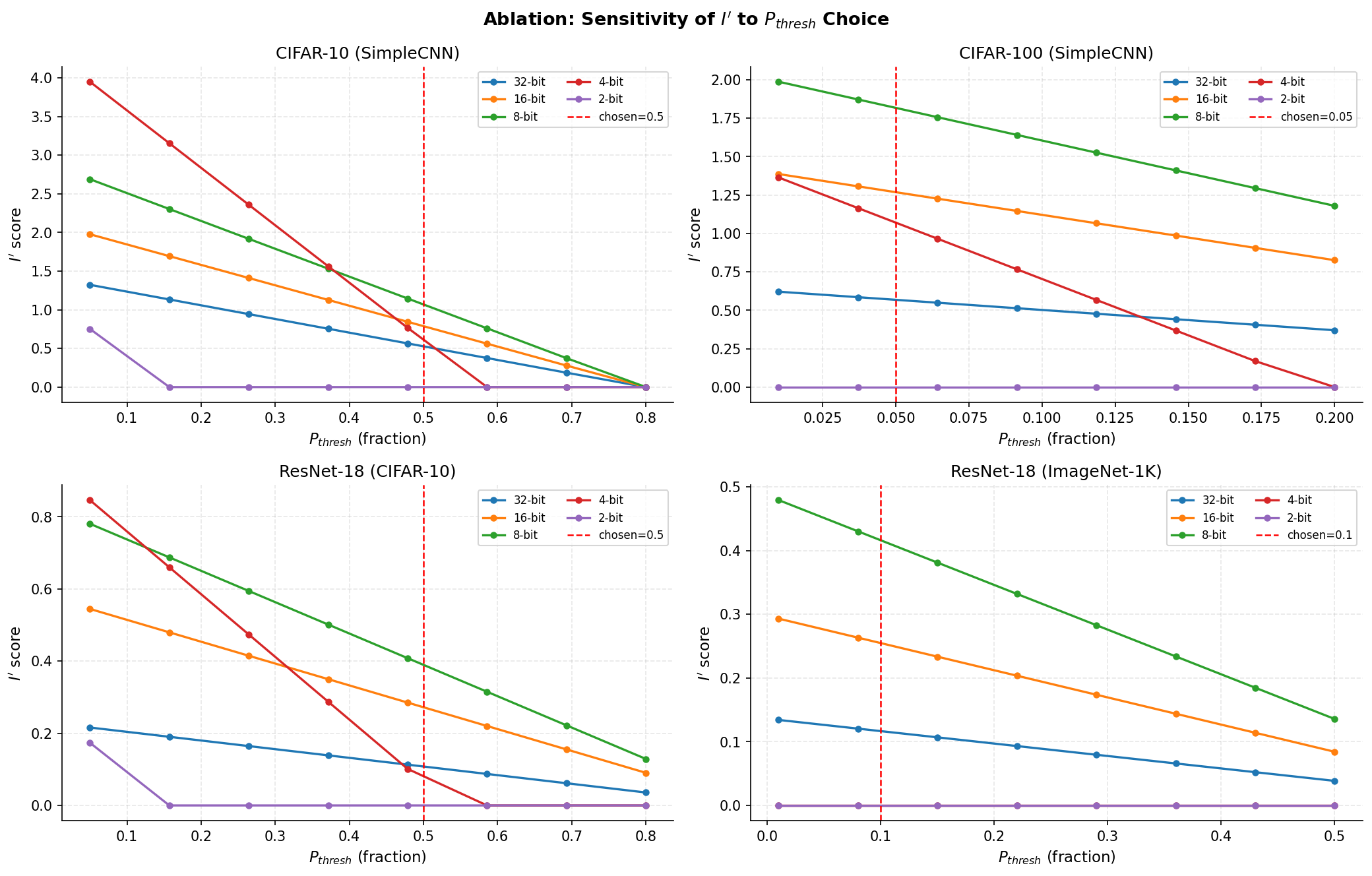}
\caption{Ablation Analysis of $P_{thresh}$: Impact of accuracy thresholds on peak efficiency.}
\label{fig:ablation_analysis}
\end{figure}

\begin{figure}[H]
\centering
\includegraphics[width=0.8 \columnwidth]{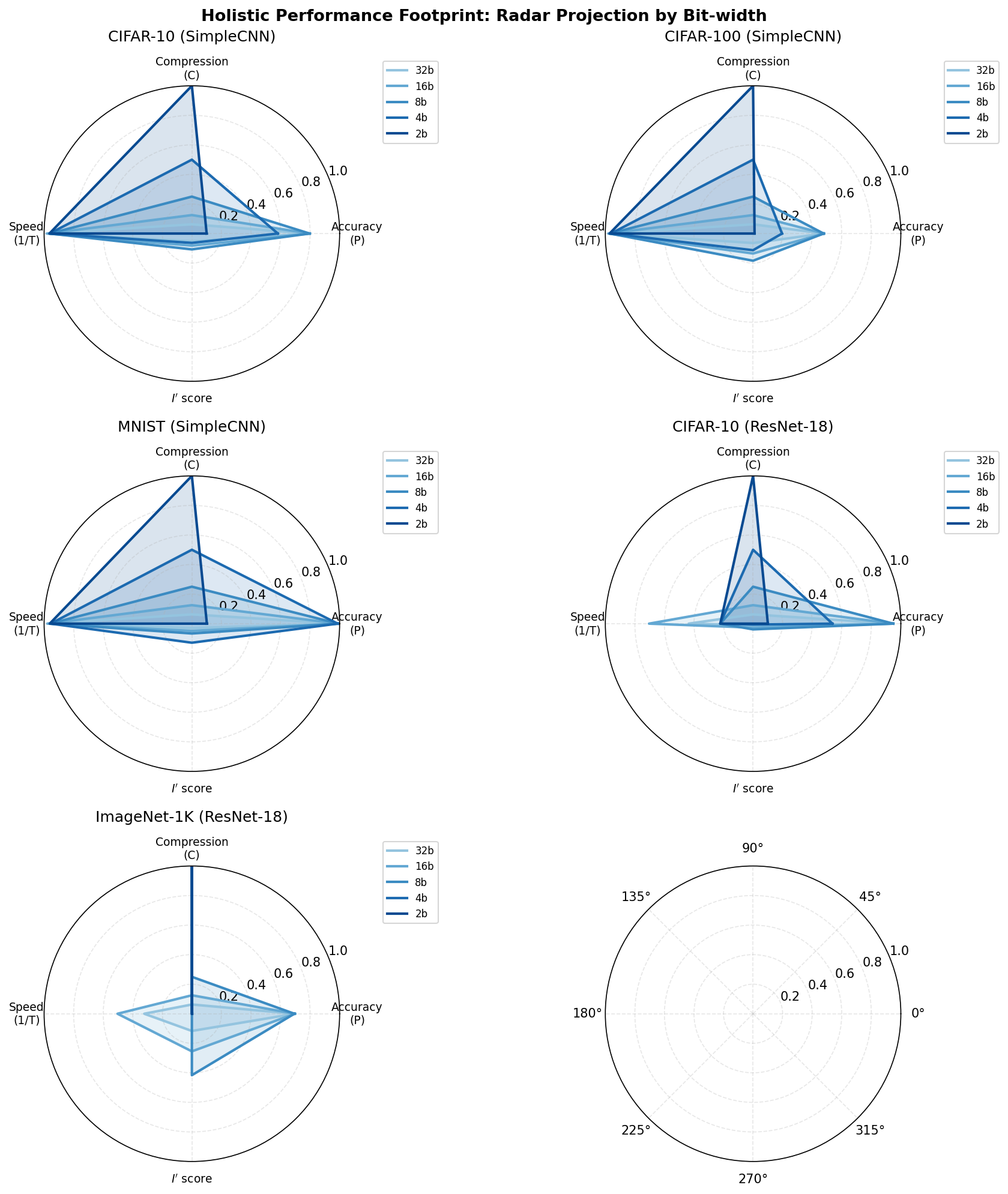}
\caption{Holistic Performance Footprint: Radar visualization of QuIDE components.}
\label{fig:dataset_comparison_radar}
\end{figure}

\begin{figure}[H]
\centering
\includegraphics[width=0.7\columnwidth]{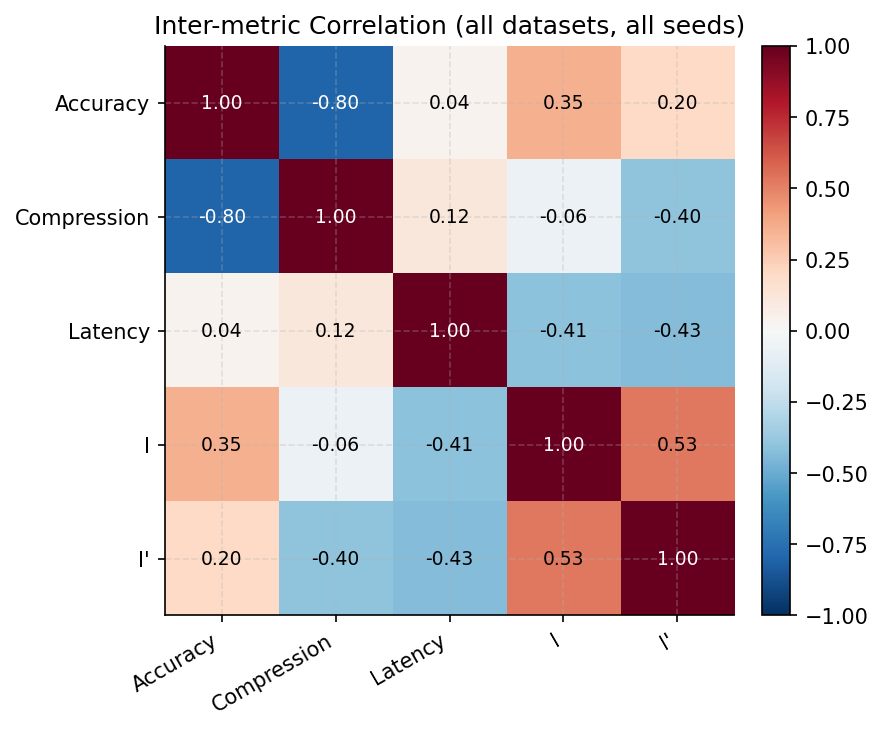}
\caption{Correlation Analysis: Inter-dependency heatmap visualizing stochastic coupling.}
\label{fig:metric_heatmap}
\end{figure}

\textbf{Visualization Suite.} The holistic footprint of QuIDE components is further explored through radar projections and correlation heatmaps. Figure \ref{fig:ablation_analysis} demonstrates the adaptability of the Intelligence Index to heterogeneous deployment requirements, revealing how the calibration of the accuracy threshold ($P_{thresh}$) shifts the efficiency peak between high-throughput and safety-critical regimes. The radar projection in Figure \ref{fig:dataset_comparison_radar} offers a geometric intuition for model capacity, where the inscribed area correlates with the scalar Intelligence Index score. The complex interplay between constituent metrics is further elucidated by the inter-dependency heatmap in Figure \ref{fig:metric_heatmap}, which highlights the stochastic coupling between precision transitions.

\begin{figure}[H]
\centering
\includegraphics[width=0.8 \columnwidth]{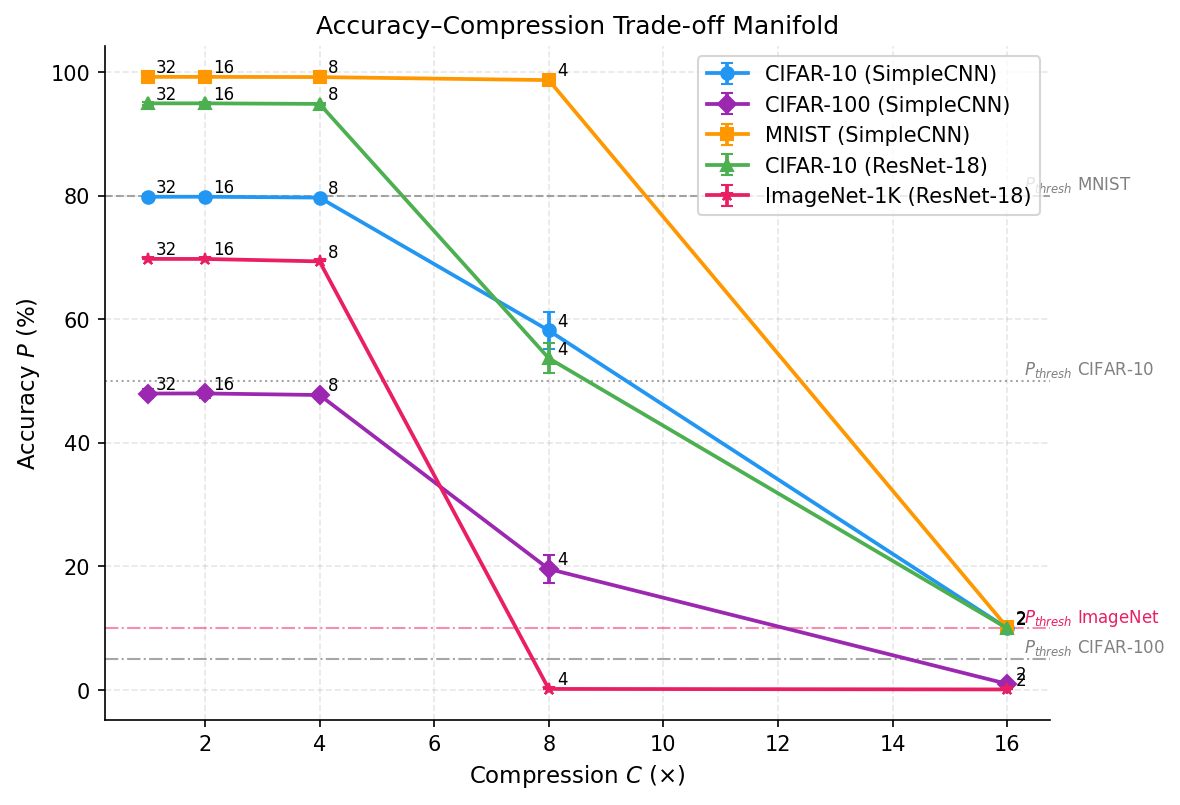}
\caption{The Accuracy-Compression Boundary: Visualization of the trade-off manifold.}
\label{fig:accuracy_vs_compression}
\end{figure}

\textbf{Relative Metric Rankings.} We evaluate the consistency of the Intelligence Index against alternative composite metrics by ranking the relative performance of the quantized model ensemble, as detailed in Table \ref{tab:metric_rankings}. The Accuracy-Compression Product ($C \times P$) mirrors the trajectory of the raw Intelligence Index, failing to penalize catastrophic accuracy loss. Conversely, Accuracy per Log-Latency ($P/\log_2(T+1)$) favors high-precision configurations, failing to reward efficiency gains at intermediate bit-widths. The refined index $I'$ provides a balanced, dataset-dependent ranking, proving its utility as a deterministic guide for optimal bit-width selection.

\begin{table}[ht]
\centering
\caption{Bit-width Priority Ranking: rank 1 = highest metric score. Bold marks the metric's top-ranked bit-width. $I$ and $ACP$ agree where no accuracy collapse occurs; $I'$ corrects them near collapse boundaries. $ALS = P/\log_2(T{+}1)$ consistently favours low-latency high-precision configurations.}
\label{tab:metric_rankings}
\resizebox{0.9\columnwidth}{!}{%
\begin{tabular}{lccccc}
\toprule
\textbf{Dataset} & \textbf{Bit-width} & \textbf{$I$ Rank} & \textbf{$I'$ Rank} & \textbf{$C{\times}P$ Rank} & \textbf{$P/\log_2(T{+}1)$ Rank} \\
\midrule
CIFAR-10 & 32 & 5 & 4 & 5 & \textbf{1} \\
CIFAR-10 & 16 & 3 & 2 & 4 & 2 \\
CIFAR-10 & 8  & 2 & \textbf{1} & 2 & 3 \\
CIFAR-10 & 4  & \textbf{1} & 3 & \textbf{1} & 4 \\
CIFAR-10 & 2  & 4 & 5 & 3 & 5 \\
\midrule
CIFAR-100 & 32 & 4 & 4 & 4 & 2 \\
CIFAR-100 & 16 & 3 & 2 & 3 & \textbf{1} \\
CIFAR-100 & 8  & \textbf{1} & \textbf{1} & \textbf{1} & 3 \\
CIFAR-100 & 4  & 2 & 3 & 2 & 4 \\
CIFAR-100 & 2  & 5 & 5 & 5 & 5 \\
\midrule
MNIST & 32 & 4 & 4 & 5 & \textbf{1} \\
MNIST & 16 & 3 & 3 & 3 & 2 \\
MNIST & 8  & 2 & 2 & 2 & 3 \\
MNIST & 4  & \textbf{1} & \textbf{1} & \textbf{1} & 4 \\
MNIST & 2  & 5 & 5 & 4 & 5 \\
\bottomrule
\end{tabular}
}
\end{table}

\section{Genetic Algorithm for Mixed-Precision Search (QuIDE-GA)}
\label{sec:appendix_ga}

To ensure full reproducibility of the adversarial benchmark presented in Section 5, we detail the complete evolutionary trajectory governed by the QuIDE fitness function. The Mixed-Precision Search (MPS) operates over the discrete configuration manifold $\mathcal{A} = \{16, 8, 4, 2\}^L$, where $L=4$ represents the architectural depth of the CNN constraint environment.

Algorithm \ref{alg:quide_ga} defines the deterministic search procedure. It explicitly demonstrates how the theoretical Intelligence Index ($I'$) transitions from a passive ranking metric into an active, gradient-free structural loss function capable of guiding the agent past heuristic boundaries.

\begin{algorithm}[ht]
\caption{QuIDE-Guided Mixed-Precision Search}
\label{alg:quide_ga}
\begin{algorithmic}[1]
\REQUIRE Full-Precision Model $\mathcal{M}_{FP}$, Search Manifold $\mathcal{A}$, Latency Penalty $\log_2(T+1)$, Accuracy Threshold $P_{thresh}$
\REQUIRE Iterations $G=30$, Population Size $N=20$, Mutation Probability $\mu=0.15$
\STATE \textbf{Initialize} population $\mathcal{P} \leftarrow \{c_1, c_2, \dots, c_N \}$, where each underlying topology $c_i \sim \mathcal{A}$ randomly
\FOR{$g = 1$ \textbf{to} $G$}
    \STATE $\mathcal{S} \leftarrow \emptyset$ \COMMENT{Fitness buffer}
    \FOR{\textbf{each} topology configuration $c_i \in \mathcal{P}$}
        \STATE Obtain discrete quantized constraint networks $\mathcal{M}^{(c_i)}$
        \STATE \textbf{Calculate} Geometric Compression: $C(c_i) = 32 / \text{mean}(c_i)$
        \STATE \textbf{Evaluate} Task Accuracy: $P(c_i)$ on Target Benchmark
        \STATE \textbf{Measure} Edge-Emulated Hardware Latency: $T(c_i)$
        \STATE \textbf{Compute Fitness}: $F(c_i) = \frac{C(c_i) \times \max(P(c_i) - P_{thresh}, 0)}{\log_2(T(c_i) + 1)}$
        \STATE Append $(c_i, F(c_i))$ to $\mathcal{S}$
    \ENDFOR
    \STATE Sort $\mathcal{P}$ descending by intelligence scores in $\mathcal{S}$
    \STATE $\mathcal{P}_{next} \leftarrow \text{Top}(K=5)$ configurations \COMMENT{Strict Elitism Preservation}
    \WHILE{$|\mathcal{P}_{next}| < N$}
        \STATE Select structural parents $p_1, p_2 \propto F(c)$ using Roulette Wheel
        \STATE Crossover $c_{child} \leftarrow \text{UniformCrossover}(p_1, p_2)$
        \IF{$\text{rand}() < \mu$}
            \STATE $c_{child} \leftarrow \text{RandomMutation}(c_{child}, \mathcal{A})$ \COMMENT{Jump local minima}
        \ENDIF
        \STATE $\mathcal{P}_{next} \leftarrow \mathcal{P}_{next} \cup \{c_{child}\}$
    \ENDWHILE
    \STATE \textbf{Update Environment}: $\mathcal{P} \leftarrow \mathcal{P}_{next}$
\ENDFOR
\RETURN $c_{best} = \arg\max_{c \in \mathcal{P}} F(c)$ \COMMENT{Yields optimal heterogeneous topology (e.g. 8-4-4-2)}
\end{algorithmic}
\end{algorithm}

\end{document}